%% file: main.tex
\documentclass[10pt,twocolumn,letterpaper]{article}

\usepackage[pagenumbers]{cvpr} %

\usepackage{graphicx}
\usepackage{amsmath}
\usepackage{amssymb}
\usepackage{booktabs}

\usepackage[pagebackref,breaklinks,colorlinks]{hyperref}

\usepackage[capitalize]{cleveref}
\crefname{section}{Sec.}{Secs.}
\Crefname{section}{Section}{Sections}
\Crefname{table}{Table}{Tables}
\crefname{table}{Tab.}{Tabs.}

\input{preamble}
\begin{document}
\input{sec/0_metadata}
\maketitle
\input{sec/0_abstract}
\input{sec/1_introduction}

\input{sec/2_related}

\input{sec/3_method}

\input{sec/4_results}

\input{sec/5_conclusions}

\input{sec/6_acknowledgements}

{
    \small
    \bibliographystyle{ieee_fullname}
    \bibliography{macros,main}
}

\clearpage

\input{sec/X_supplementary}

\end{document}

%% file: preamble.tex
\usepackage{overpic}
\usepackage{enumitem} %
\usepackage{overpic} %
\usepackage{color}
\usepackage{multirow}

\definecolor{turquoise}{cmyk}{0.65,0,0.1,0.3}
\definecolor{purple}{rgb}{0.65,0,0.65}
\definecolor{dark_green}{rgb}{0, 0.5, 0}
\definecolor{orange}{rgb}{0.85, 0.5, 0.1}
\definecolor{red}{rgb}{0.8, 0.2, 0.2}
\definecolor{darkred}{rgb}{0.6, 0.1, 0.05}
\definecolor{blueish}{rgb}{0.0, 0.3, .6}
\definecolor{light_gray}{rgb}{0.7, 0.7, .7}
\definecolor{pink}{rgb}{1, 0, 1}
\definecolor{greyblue}{rgb}{0.25, 0.25, 1}
\definecolor{orange_bright}{rgb}{1, 0.5, 0.1}

\DeclareMathOperator*{\argmin}{arg\,min}

\newcommand{\loss}[1]{\mathcal{L}_\text{#1}}
\newcommand{\hparam}[1]{\lambda_\text{#1}}

\newcommand{\quant}{\mathsf{q}}
\newcommand{\MaskNet}{\mathcal{M}}
\newcommand{\GenNet}{\mathcal{G}}
\newcommand{\Encoder}{\mathcal{E}}
\newcommand{\Decoder}{\mathcal{D}}
\newcommand{\Codebook}{\mathcal{Z}}
\newcommand{\Control}{\mathcal{T}}
\newcommand{\Discrim}{\mathcal{C}}
\newcommand{\VGG}{\mathcal{V}}
\newcommand{\mbf}{\mathbf{f}}
\newcommand{\bm}{\mathbf{m}}
\newcommand{\bz}{\mathbf{z}}
\newcommand{\btheta}{\boldsymbol{\theta}}
\newcommand{\IR}{\mathbb{R}}
\newcommand{\IE}{\mathbb{E}}
\newcommand{\roughmask}{\boldsymbol{\mu}}

\newcommand{\Figure}[1]{Figure~\ref{fig:#1}}

\newcommand{\Table}[1]{Table~\ref{tab:#1}}

\newcommand{\Eq}[1]{Eq.~\eqref{eq:#1}}
\newcommand{\Equation}[1]{Equation~\eqref{eq:#1}}

\newcommand{\Section}[1]{Section~\ref{sec:#1}}

\usepackage{blindtext}

\def \customparskip {0.5em}
\renewcommand{\paragraph}[1]{\vspace{\customparskip}\noindent\textbf{#1}.}
\renewcommand{\subparagraph}[1]{\vspace{\customparskip}\noindent--- #1.}

\newcommand{\supplementary}{\textit{Supp. Mat.}\xspace}

%% file: sec/0_metadata.tex
\title{Layered Controllable Video Generation}

\author{%
    Jiahui Huang$^{1,2}$\qquad
    Yuhe Jin$^{1}$\qquad
    Kwang Moo Yi$^{1}$\qquad
    Leonid Sigal$^{1,2,3,4}$\\[.2in]
    $^1$University of British Columbia\qquad 
    $^2$Vector Institute for AI \qquad \\
    $^3$CIFAR AI Chair \qquad
    $^4$NSERC CRC Chair \qquad
    \\[8pt]
    {\tt\small \href{https://gabriel-huang.github.io/layered_controllable_video_generation/}{layered-controllable-video-generation.github.io}}}

%% file: sec/0_abstract.tex
\begin{abstract}
We introduce layered controllable video generation, where we, without any supervision, decompose the initial frame of a video into foreground and background layers, with which the user can control the video generation process by simply manipulating the foreground mask.
The key challenges are the unsupervised foreground-background separation, which is ambiguous, and ability to anticipate user manipulations with access to only raw video sequences.
We address these challenges by proposing a two-stage learning procedure.
In the first stage, with the rich set of losses and dynamic foreground size prior, we learn how to separate the frame into foreground and background layers and, conditioned on these layers, how to generate the next frame using VQ-VAE generator.
In the second stage, we fine-tune this network to anticipate edits to the mask, 
by fitting (parameterized) control to the mask from future frame. 
We demonstrate the effectiveness of this learning and the more granular control mechanism, while illustrating state-of-the-art performance on two benchmark datasets.
\end{abstract}

%% file: sec/1_introduction.tex
\section{Introduction}
\label{sec:intro}

Advances in deep generative models have led to impressive results in image and video synthesis.
Typical forms of such models, including Variational Autoencoder (VAE)~\cite{KingmaVAE}, Generative Adversarial (GAN)~\cite{GoodfellowGAN} and recurrent (RNN)~\cite{OordPixelRNN} formulations, can produce complex and highly realistic content.
However, synthesis of realistic images/videos, without the ability to control the depicted content in them, has limited practical utility. 
This has led to a variety of conditional generative tasks and formulations.

In the image domain, both coarse- (\eg, sentence \cite{Zhang_ICCV17}) and fine-level (\eg, layout \cite{Zhao_CVPR19} and instance attribute \cite{Frolov_GCP21}) control signals have been explored.
The progress on the video side, on the other hand, has generally been more modest, in part due to an added challenge of synthesizing temporally coherent content. 
{\em Future frame prediction}~\cite{Denton_ICML2018,finn_NIPS16,franceschi_ICML20,Kumar_ICLR20,mathieu_ICLR16,vondrick_ICLR17,vondrick_NIPS17,vondrick_CVPR17,Weissenborn_ICLR20} conditions the future generated frames on one (or a couple) seed frame(s).
But this provides very limited control as the object(s), or person(s), depicted in the conditioned frame can move in a multitude of ways, particularly as predictions are made longer into the future.
To address this, a number of methods condition future frames on the action \cite{He_ECCV18,Tulyakov_CVPR18} and object \cite{Nawhal_ECCV20} label.
Still, they only provide very coarse global video-level control. 

\begin{figure}
    \centering
    \vspace{2.5em}
    \includegraphics[width=0.85\linewidth]{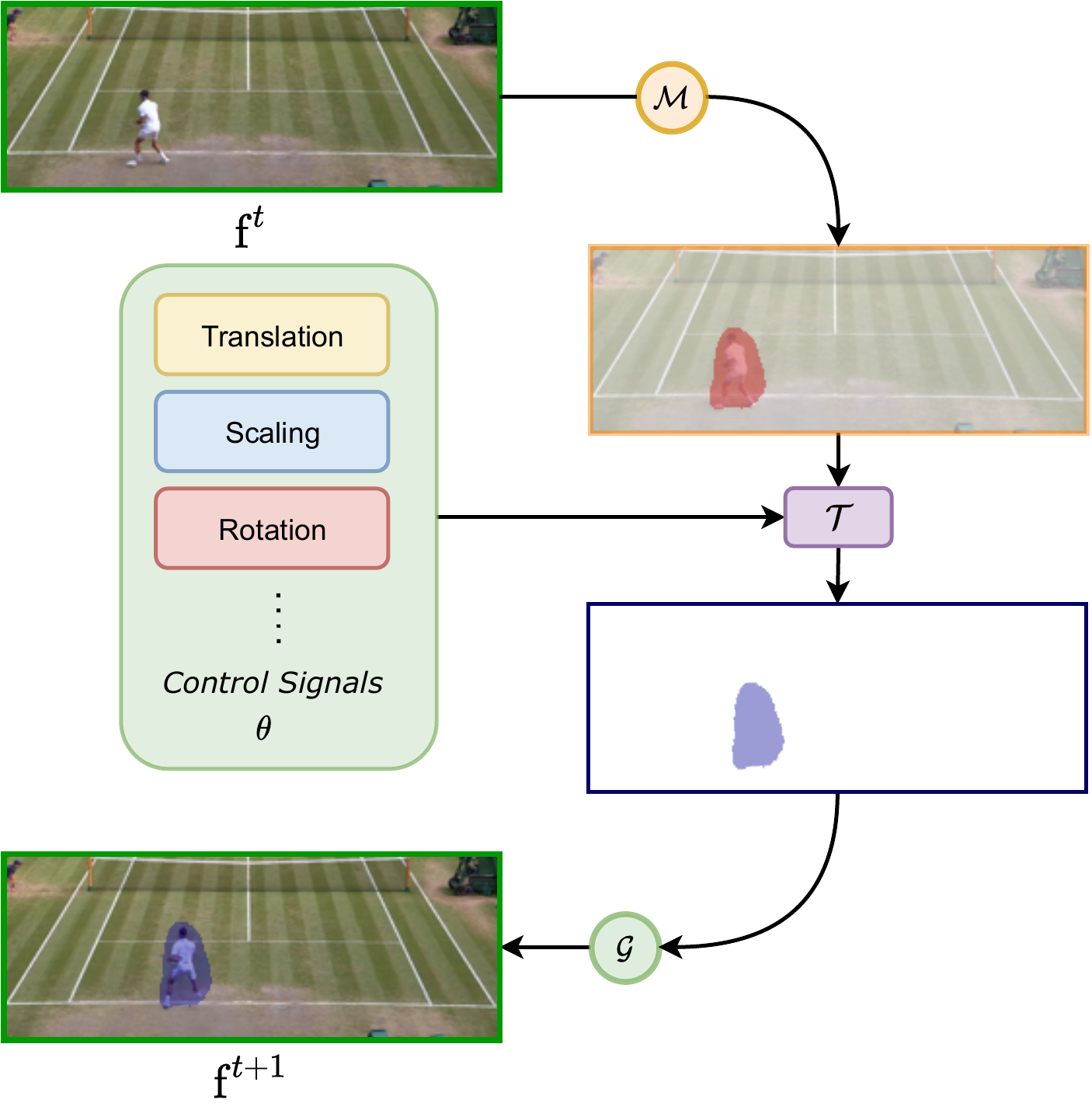}
    \caption{{\bf Layered Controllable Video Generation.} 
    Illustration of the proposed task, where a frame at time $t$ is first decomposed into a foreground/background layers, using the learned mask network $\mathcal{M}$, and then the user is allowed to modify this mask with control signals $\theta$ (\eg, by shifting it by $\Delta_t$) to control the next generated frame realized by generator $\mathcal{G}$. The foreground source and target mask are illustrated in red and blue.}
    \label{fig:front_fig}
\end{figure}

More recent approaches focus on the ability to control the video content on-the-fly at the frame-level.
Typically, these methods are formulated as conditional auto-regressive (or recurrent) models that generate one frame at a time, conditioned, for example, on the discrete action label~\cite{Kim_CVPR20} or keypoint-based human pose specification~\cite{Villegas_ICML17,Walker_ICCV17,Yang_ECCV18} (\eg, obtained from a target video source \cite{Zablotskaia_BMVC19}).
Such methods, however, require dense per-frame annotation of actions or poses at training time, which are costly to obtain and make it challenging to employ such approaches in realistic environments.
The task of {\em playable video generation}~\cite{playable}, has been introduced to address these limitations.
In playable video generation, the discrete action space is {\em discovered} in an unsupervised manner and can then be used as a conditioning signal in an auto-regressive probabilistic generative model.
While obtaining impressive results, with no intermediate supervision and allowing frame-level control over the generative process,~\cite{playable} is inherently limited to a single subject and a small set of discrete action controls. 

Thus, in this work, we aim to allow richer and more granular control over the generated video content, while similarly requiring no supervision of any kind -- \ie, having only raw videos as training input, as in~\cite{playable}.
To do so, we make an observation (inspired by early works in vision \cite{adelson_PAMI94,jojic_CVPR01,kumar_IJCV08}) that a video can be effectively decomposed into foreground / background layers.
The background layer corresponds to static (or slowly changing) parts of the scene. 
The foreground layer, on the other hand, evolves as the dominant objects move. %
Importantly, the foreground mask itself contains position, size and shape information necessary to both characterize and render the moving objects appropriately.
Hence, this foreground mask can in itself be leveraged as a useful level of abstraction that allows both intuitive and simple control over the generative process. 

With the above intuition, we therefore propose an approach that automatically learns to segment video frames into foreground and background, and at the same time 
generates future frames conditioned on the segmentation and the past frame, iteratively.
To allow on-the-fly control of video generation, we expose the foreground mask to the user for manipulation -- \eg, translations of the mask leads to corresponding in-plane motion of the object, resizing it would lead to the depth away/towards motion, and changes in the shape of the mask can control the object pose.

From the technical perspective, the challenge of formulating such a method is two fold. 
First, unsupervised foreground/background layer segmentation is highly ambiguous, particularly if the background is not assumed to be static and the foreground motions can be small.
Second, user input needs to be anticipated to ensure model learns how to {\em react} to changes in the mask, without explicit access to such information.
To this end, we propose a two-stage learning procedure.
In the first stage, the network learns how to perform foreground/background separation and, conditioned on this layered representation, future frame prediction.
Specifically, we introduce a set of sophisticated losses and a dynamic prior to learn how to predict a foreground mask and leverage VQ-VAE~\cite{VQVAE} to predict foreground and background latent content which is then fused and decoded to the next frame.
In the second stage, we simulate user input and fine-tune the generative model such that this user input can be appropriately handled. 

\vspace{0.1in}
\noindent
{\bf Contributions:} Our contributions are multi-fold.
\vspace{-.5em}
\begin{itemize}[leftmargin=*]
    \item 
    From raw video data, our model learns to generate foreground/background separation masks in an unsupervised manner.
    We then leverage the foreground layer as a flexible (parametric) user control mechanism for the generative process.
    This provides both richer and more intuitive control compared to action vectors \cite{playable} or sparse trajectories \cite{HaoCVPR18}.
    
    \item To effectively train our model we introduce two-stage training: the first stage tasked with learning how to separate layers and perform future frame prediction; the second, to adopting and anticipating user control. 
    \item To prevent over-/under-segmentation, we regularize layer separation with sparsity loss and dynamic mask size prior.
    \item Finally, we validate our approach on multiple datasets and show that we are able to generate state-of-the-art results and, at the same time, allow higher level of control over the generated content without any supervision.
\vspace{-.25em}
\end{itemize}

%% file: sec/2_related.tex
\section{Related works}
\label{sec:related}
\vspace{-\customparskip}
\paragraph{Video generation}
Early video generation techniques proposed to generate a video as a whole. Most of these convert a noise vector, sampled from simple distribution or a prior, to a video, using a GAN \cite{vondrick_NIPS17,Acharya_2018,Wang_CVPR2020} or VAE \cite{Babaeizadeh_2018,Denton_ICML2018} formulation. More recent architectures leveraged transformer-based formulations \cite{Yan_2021,Neimark_2021,Rakhimov_2021,Weissenborn_ICLR20} that have generally resulted in higher quality video outputs. As an alternative to 3D (transposed) convolution techniques, that generate all frames at once, recurrent auto-regressive variants \cite{Kalchbrenner_2017} have also been explored. 
While these early works largely focused on the video quality and resolution, more recently, the focus has shifted to conditioned or controlled video generation.

\input{fig/pipeline_item}

\paragraph{Video generation with global control}
Future frame prediction considers the task of generating a video conditioned on a few starting seed frames. Early approaches to future prediction employed deterministic models \cite{finn_NIPS16,mathieu_ICLR16,vondrick_ICLR17,Wichers2018HierarchicalLV} that failed to model uncertainty in the future induced by variability of unfolding events. To overcome this limitation, later methods, based on VAE \cite{Babaeizadeh_2018,Denton_ICML2018}, GAN \cite{Lee_2018,Kwon_CVPR2019}, and probabilistic formulations \cite{Xue2019VisualDS}, attempted to introduce real-world stochasticity into the generative process. Action label conditioning, in combination with seed frames or not, where a video sequence is generated conditioned on an input action label \cite{Kim_Neurips2019,Wang_WACV2020} is also popular; some such approaches leverage disentangled factored representations (\eg, of subject identity and action \cite{He_ECCV18}). Other types of global conditioning signals include action-object tuples \cite{Nawhal_ECCV20}. However, these methods, collectively, require action annotations for training, and, more importantly, do not allow control at the frame-level. 

\paragraph{Video generation with frame-level control}
More granular control, at the frame-level, has also been explored. Pose-guided generative models first generate a sequence sparse \cite{Walker_ICCV17,Villegas_ICML17,Yang_ECCV18} or dense \cite{Zablotskaia_BMVC19} keypoint human poses, either predictively \cite{Villegas_ICML17,Walker_ICCV17} or from a source video \cite{Zablotskaia_BMVC19}, and then use these for conditional generation of respective video frames. However, these methods are only applicable to videos of human subjects and require either pose annotations or a pre-trained pose detector. Alternatively, individual frames can be conditioned on action labels \cite{Kim_CVPR20,Chiappa_2017,Oh_NIPS2015}. However, these methods require dense frame-level annotations, which are only available in limited environments such as video games. Closest to our work, Menapace \etal proposed Playable Video Generation \cite{playable}, which, in an unsupervised manner, discovers semantically consistent actions meanwhile generating the next frame conditioned on the past frames and an input action, thus providing user control to the generation process. However, their method is limited to a small set of discrete action controls and explicitly assumes a single moving object. In contrast, our method allows richer and more granular control, and can be used to generate, and control, videos with multiple objects.

\paragraph{Video generation with pixel-level control} 
It is worth mentioning that some prior works attempted to use dense semantic segmentation \cite{Pan_CVPR2019,WangNIPS19} and sparse trajectories \cite{HaoCVPR18} to control video generation. While such approaches allow granular control at a pixel-level, these representations are incredibly difficult for a user to produce or modify. We also use a form of (foreground) segmentation for control, however, it is unsupervised, class-agnostic, and can be easily controlled either parametrically or non-parametrically.

\paragraph{Unsupervised object segmentation}
Layered representations have long history in computer vision \cite{adelson_PAMI94,jojic_CVPR01,kumar_IJCV08}, and are supported by neuroscience evidence \cite{WebsterBrain}.
Traditional techniques for this rely on feature clustering \cite{Achanta_2012,Alexe_2010,Hochbaum_2009} and statistical background modeling and subtraction \cite{BouwmansBGBook,ElgammalECCV00,StaufferCVPR99}.
Such techniques work best for videos where the background is (mostly) static, lighting fixed and the foreground is fast moving; we refer readers to \cite{Changedetection} for an extensive analysis and discussion.
More recent techniques have focused on generative formulations for the task.
In particular, Bielski \etal proposed the PerturbGAN \cite{Bielski_NeurIPS2019}, their model generates the foreground and background layers separately, and uses a perturbation strategy to enforce the generation of semantically meaningful masks.
Related, MarioNette \cite{SmirnovNIPS21} learns to decompose scenes into a background and a learned dictionary of sprites. Other approaches focus on separation of videos into natural layers (\eg, to factorize secondary effects such as shadows and reflections \cite{alayrac2019visual}) and to control which layer to attend to \cite{AlayracICCV2019}.
Similar to~\cite{vondrick_NIPS17}, we decompose frames and separately model foreground/background content that is then composed/fused together to produce video.
However, unlike~\cite{vondrick_NIPS17} and others, we allow the user to have frame-level control over the foreground mask using both intuitive parametric and nonparametric controls.

%% file: fig/pipeline_item.tex
\begin{figure*}[t!]
    \centering
    \includegraphics[width=0.85\linewidth]{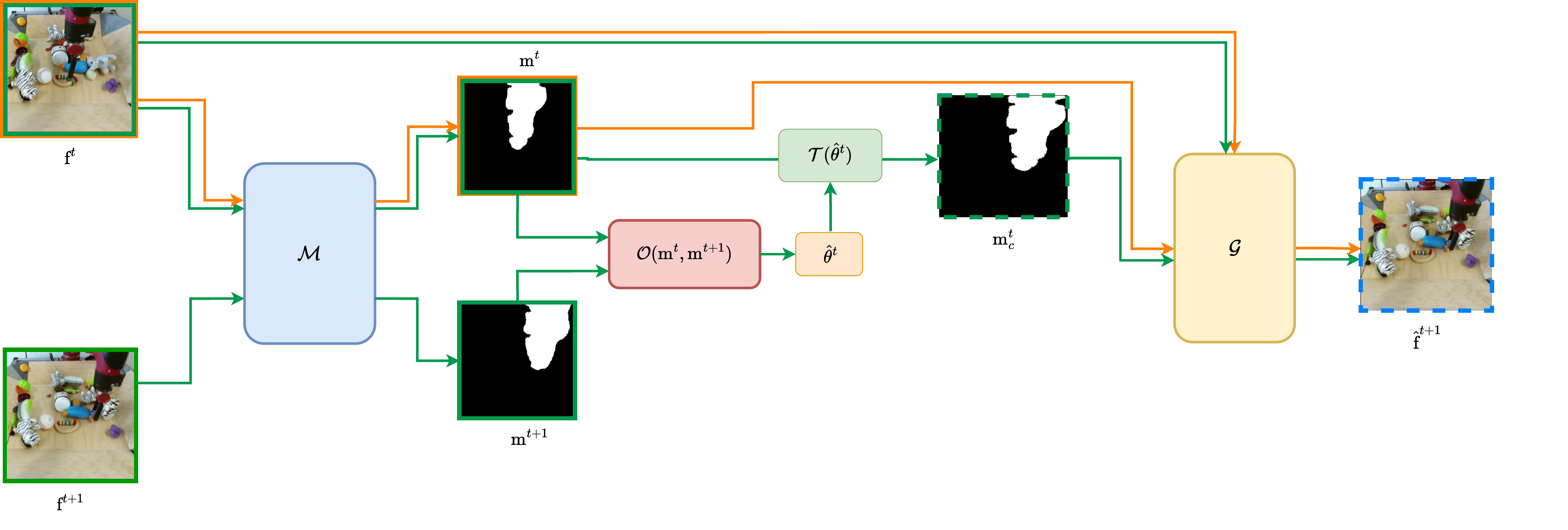}
    \caption{
    {\bf Illustration of the Proposed Two-stage Training.} %
    The flow of Stage I is represented in {\color{orange_bright}orange} and Stage II in  {\color{dark_green}green}. $\mathcal{M}$ denotes the {\em mask network}, which estimates the foreground/background mask, and $\mathcal{G}$ denotes the {\em generator network} that takes the current frame and a mask to produce the next frame. 
    $\mathcal{O}(\cdot)$ is the optimization procedure described in Eq.(\ref{eq:optimization}). $\mathcal{T}(\cdot)$ is a differentiable function that transforms a mask to a target shape using {\em user control signal} $\theta^t$.
    }
    \label{fig:pipeline} 
    
\end{figure*}

%% file: sec/3_method.tex
\section{Method}
\label{sec:method}

We first discuss our framework in detail then our two-stage training strategy.

\subsection{Framework}

\vspace{-\customparskip}
\paragraph{Inference}
For easy explanation, we first explain the inference flow of our method. 
As illustrated in \Figure{pipeline}~(b), our method is composed of mainly two components: the mask network and the generator network.
The \textit{mask network} $\MaskNet$ takes as input an image frame $\mbf^t\in\IR^{3 \times W\times H}$ at time $t$ and outputs a mask:
\begin{equation}
    \bm^t=\MaskNet\left(\mbf^t\right)
    ,
    \label{eq:masknet}
\end{equation}
where $\bm^t \in \IR^{W\times H}$,
segmenting the foreground layer and the background layer.
Then, our \textit{generator network} $\GenNet$ takes the mask and the frame as input to generate the next frame:
\begin{equation}
    \hat{\mbf}^{t+1}=\GenNet\left(\mbf^t, \bm^t\right)
    .
    \label{eq:gennet}
\end{equation}
More specifically, for the generator, we employ a VQ-VAE framework \cite{VQVAE} for its high-quality image generation.
Hence, as shown in \Figure{G}, the generator $\GenNet$ can be written as a composite function of an encoder $\Encoder$, a decoder $\Decoder$, and a learnable discrete code book $\Codebook = \{\bz_k\}^K_{k=1}\subset \IR^{n_z}$, where $n_z$ is the dimension of each code, we encode the foreground layer and the background layer separately in our pipeline, and then merge them in the latent space before feeding them into the decoder:
\begin{equation}
\begin{split}
    \bz_{fg}^t &=\Encoder\left(\mbf^t \odot \bm^t\right) \\
    \bz_{bg}^t &=\Encoder\left(\mbf^t \odot (1-\bm^t)\right) \\
    \hat{\mbf}^{t+1} &=\Decoder\left(
        \quant\left(\bz_{fg}^t\right)
        +
        \quant\left(\bz_{bg}^t\right)
    \right)
\end{split}
    \;,
    \label{eq:vqvae}
\end{equation}
where $\odot$ is the element-wise product and $\quant$ is the element-wise quantization function defined as:
\begin{equation}
    \quant(\bz)_{ij} := 
        \underset{\bz_k \in \Codebook}{\arg\min } 
        \left\| \bz_{ij}-\bz_k \right\| 
    ,
\end{equation}
where $i$ and $j$ are the row and column indices.
$\GenNet$ and $\MaskNet$ are then used to generate the video via auto-regression.

\input{fig/G_item}

\paragraph{Control}
With the above inference pipeline, we enable on-the-fly user control by modifying the mask to create $\bm^t_c$ and using it in place of $\bm^t$ in \Eq{gennet}.
Mathematically, we write
\begin{equation}
    \bm^t_c = \Control\left(\bm^t, \btheta^t\right)
    ,
    \label{eq:control}
\end{equation}
where $\Control$ is the mask controlling operation parameterized by $\btheta^t$.
In the crudest form, the controlling operation $\Control$ can simply be shifting the mask $\bm^t$ by $\Delta x$ and $\Delta y$  in horizontal and vertical directions, respectively, or for example, be an affine transformation.
Both can be implemented differentiably as a simple parametric coordinate transformation on $\bm^t$~\cite{Jadeberg15,Jiang19}, allowing differentiable control of the mask.
We utilize this the differentiabilty later in \Section{results} to find the ``ground-truth'' control signal for easy comparison with existing works. 
Other forms of control are also possible, including more granular non-parametric, and non-differentiable, direct manipulation of the mask.

\paragraph{Training}
Given the inference pipeline, we now wish to train the three networks $\MaskNet$, $\Encoder$ and $\Decoder$, \textit{without} any supervision. 
In order to do this, we introduce two main ideas:\\
\vspace{-1.5em}
\begin{itemize}[leftmargin=*]
    \setlength\itemsep{-.3em}
    \item we can treat the mask created by $\MaskNet$ as a latent variable, which with \textit{proper regularization} during training,
    can be discovered automatically (\Section{phaseI});
    \item we can simulate the user input by considering the mask at time $t$ and $t+1$ (\Section{phaseII}).
\end{itemize}
We detail them in the following subsections. %

\subsection{Initial training for mask-based generation}
\label{sec:phaseI}
We first train our method
with the focus of generating high-quality foreground-background segmentation masks, without considering introducing controlability to the generation process. 
Doing so requires two main objectives: (1) high-quality estimation of the next image frame $\mbf^{t+1}$; and (2) proper regularization of the generated mask.
Hence, if we write the two objectives as loss terms $\loss{img}$ and $\loss{mask}$, respectively, the total loss for the first stage training of our method $\loss{total}$ can be written as:
\begin{equation}
    \loss{total}=\loss{img}+\loss{mask}
    .
    \label{eq:loss_total}
\end{equation}
We detail each loss term in the following.

\paragraph{Learning to predict the next frame}
To train our method to estimate the next frame, we follow the VQ-VAE\cite{VQVAE} framework with a Generative Adversarial Network (GAN) loss to encourage realism\cite{taming}.
We therefore write
\begin{equation}
    \loss{img}=\hparam{VQ}\loss{VQ} + \hparam{GAN}\loss{GAN} + \hparam{percept}\loss{percept}
    ,
\end{equation}
where $\hparam{VQ}$, $\hparam{GAN}$, $\hparam{percept}$ are hyperparameters controlling the influence of each loss term.

\subparagraph{$\loss{VQ}$}
Following the original VQ-VAE formulation we write
\begin{equation}
\begin{split}
    \loss{VQ} 
    = \left\|\mbf^{t+1} - \hat{\mbf}^{t+1} \right\|_1
    & + \left\| \text{sg}\left[\Encoder(\mbf^t)\right] - \bz_\mathsf{q} \right\|^2_2 \\
    & + \left\| \text{sg}\left[\bz_\mathsf{q}\right] - \Encoder(\mbf^t) \right\|^2_2    
\end{split}
    ,
\end{equation}
where $\mbf^{t+1}$ and $\hat{\mbf}^{t+1}$ are true and estimated next frame, $\text{sg}[\cdot]$ denotes the stop gradient operation, and $\bz_\mathsf{q}$ is the quantized latent variable of the VQ-VAE.
Note that we use the $\ell_1$ norm, instead of the $\ell_2$, for the reconstruction part of the loss, as we emprically found it to be more stable in training.

\subparagraph{$\loss{GAN}$}
We train a discriminator $\Discrim$ with the architecture from \cite{pix2pix} and aim to improve the generation quality.
We therefore write
\begin{equation}
    \loss{GAN} = \log\left(\Discrim\left(\mbf^{t+1}\right)\right) + \log\left(1-\Discrim\left(\hat{\mbf}^{t+1}\right)\right)
    .
\end{equation}
We note that for the hyperparameter for this loss $\hparam{GAN}$, we follow \cite{taming} and apply a dynamic weighing strategy, which stabilizes training.

\subparagraph{$\loss{percept}$}
We use a pretrained VGG-16 network~\cite{VGG} to extract deep features and compute the perceptual loss.
Denoting the deep feature extraction process as $\VGG$ we write
\begin{equation}
    \loss{percept} = \left\|\VGG(\hat{\mbf}^{t+1}) - \VGG(\mbf^{t+1}) \right\|_2
    .
\end{equation}

\paragraph{Regularizing the mask}
To train our method to generate proper masks without any supervision, we regularize the mask with three losses: 
(1) $\loss{bg}$ -- the contents of the background should not change; 
(2) $\loss{fg}$ -- there should be as least amount of foreground as possible  since classifying all pixels as foreground provides a trivial solution for $\loss{bg}$; and 
(3) $\loss{bin}$ -- the masking should be binary for effective separation.
We therefore write the mask regularization loss $\loss{mask}$ as
\begin{equation}
\label{eq:mask_loss}
    \loss{mask}=\hparam{bg}\loss{bg}+\hparam{fg}\loss{fg}+\hparam{bin}\loss{bin}
    ,
\end{equation}
where $\hparam{bg}$, $\hparam{fg}$, and $\hparam{bin}$ are the hyperparameters controlling how much each loss term affects the mask regularization.

\subparagraph{$\loss{bg}$}
We aim to ensure that the mask correctly identifies the background, \ie, non-moving parts of the scene. Hence, we simply define it as the amount of change in the masked out (background) region between two consecutive frames.
We write
\begin{equation}
\begin{split}
    \loss{bg}&=
    \left\|
        (1-\bm^t)\odot\mbf^t - (1-\bm^t)\odot\mbf^{t+1}
    \right\|_1
\end{split}
    ,
    \label{eq:loss_bg}
\end{equation}
where $\odot$ denotes the elementwise multiplication.
Note that we define this loss using the $\ell_1$ norm, as changes in the scene are not strictly restricted to the mask -- \eg shadows of moving objects can occur in the background, or other scenic changes, such as a global illumination change can happen -- and the $\ell_1$ norm leaves room for the method to incorporate these edits if necessary.

\subparagraph{$\loss{fg}$}
As mentioned earlier, $\loss{bg}$ alone, leaves room for a trivial solution---assigning $\bm^t{=}1$ results in $\loss{bg}{=}0$ regardless of the values of $\mbf^t$ and $\mbf^{t+1}$.
This could be avoided by enforcing an additional loss term that penalizes having too many foreground pixels, but a na\"ive regularization is not sufficient, as the amount of the actual foreground pixels to be edited may drastically differ from one frame to another -- \eg, robot arm moving close to the camera vs. further away.

We thus propose to regularize based on the amount of evident change between $\mbf^t$ and $\mbf^{t+1}$, approximated using simple background subtraction. 
Specifically, for a pixel index $i$, $j$, if we denote whether the pixel changed between the two consecutive frames $\mbf^t$ and $\mbf^{t+1}$ as
\begin{align}
    \roughmask^t_{ij}=
    \begin{cases}
        1, & \text{if } \left\|\mbf^{t+1}_{ij}-\mbf^{t}_{ij}\right\|_1 > \tau \\
        0, & \text{otherwise}
    \end{cases}
    ,
\end{align}
where $\tau$ is a threshold for controlling the sensitivity, we can use the average of $\roughmask_{ij}^t$ as a rough standard
for how much of the pixels should be foreground, dynamically for each consecutive frame.
We thus write
\begin{equation}
    \loss{fg} = \max\left\{0, \left\|\IE_{ij}\left[{\bm}_{ij}^t\right]-\IE_{ij}\left[{\roughmask}_{ij}^t\right]\right\|_1\right\}
    ,
    \label{eq:loss_fg}
\end{equation}
where $\IE_{ij}$ is the expectation over pixels.

We note that the balance between the hyperparameter settings related to $\loss{fg}$ and $\loss{bg}$ is important since they govern how the loss behaves---\eg a wider mask that covers all potential changes in the scene, or a tight mask that only focuses on the actual changing locations.
We empirically set the ratio between $\lambda_{bg}$ and $\lambda_{fg}$ to be 100:1, and we gradually decrease this ratio in training for faster convergence.

\subparagraph{$\loss{bin}$}
Finally, there is one last loophole for the network to cheat its way through the two mask regularizors -- by producing intermediate values in the range $[0, 1]$.
In fact, we found in our early experiments that this \textit{soft mask} allows the deep network to encode information about the image, \ie, $\mbf^{t+1}$, thus being able to fully reconstruct the original image from mask alone, and any modification on the mask -- \eg, position shifts -- results in a global appearance change of the generated image, which is undesirable.
Therefore, we encourage the mask to be binary as in \cite{Bielski_NeurIPS2019}:
\begin{equation}
    \loss{bin} = \min\{\bm^t, (1-\bm^t)\}
    .
    \label{eq:loss_bin}
\end{equation}
We then, at test time, convert these pseudo-binary masks into hard ones by thresholding with 0.5.

\subsection{Fine-tuning for controllability}
\label{sec:phaseII}
We now discuss the second stage of our training setup, where we shift our focus to imbue our method with controllability.
While the model trained in \Section{phaseI} is a generative model conditioned on the latent mask $\bm^t$, it cannot immediately be used with any arbitrary mask -- the generator $\GenNet$ would expect a mask that aligns perfectly with $\mbf^t$, whereas our user controlled mask $\Control\left(\bm^t, \btheta^t\right)$ will not.
In other words, we need a way to simulate user input, in terms of mask modifications, and incorporate it into the training.

We therefore turn our attention to the fact that the masks between two consecutive frames $\bm^t$ and $\bm^{t+1}$ are not too different, and that $\bm^{t+1}$ can be seen as user modified version of $\bm^t$ that specifies how the foreground should move, that is,
\begin{equation}
    \bm^{t+1} \approx \Control\left(\bm^t,\btheta^t\right)
    .
\end{equation}
Under this assumption, we can then find the pseudo user control at time $t$ using the following optimization procedure \cite{Jadeberg15,Jiang19}:
\begin{equation}
    \hat{\btheta}^t \equiv \argmin_{\btheta}\left\|
    \bm^{t+1} - \Control\left(\bm^t,\btheta\right)
    \right\|
    .
    \label{eq:optimization}
\end{equation}
Now, with $\hat{\btheta}^t$, we can fine-tune our network $\GenNet$ to create the next frame, which results in controllable video generation.
Denoting the binarization operation as $\lfloor\cdot\rfloor_{0.5}$, instead of \Equation{gennet}, we write
\begin{equation}
    \tilde{\mbf}^{t+1}=\GenNet\left(\mbf^t, \left\lfloor
        \Control\left(\bm^t, \hat{\btheta}^t\right)
    \right\rfloor_{0.5}\right)
    .
    \label{eq:next_frame}
\end{equation}
We then use $\tilde{\mbf}^{t+1}$ in our loss functions to fine-tune. %

One noteworthy aspect of this second stage training is that, because we binarize the mask, no gradient flows through to $\MaskNet$.
We found this to be important, as allowing \textit{any} softness in the mask resulted in the image information leaking through the mask, resulting in non-controllabilty -- \eg, shifting the mask resulted in a shift of the entire scene.
This also leads to $\loss{bg}$, $\loss{fg}$, and $\loss{bin}$ not affecting training.
While the latter two can be dropped since they are purely on how the mask network $\MaskNet$ behaves, completely dropping $\loss{bg}$ now has the danger of the generated image ignoring the mask.
Hence, we replace $\mbf^t$ in \Equation{loss_bg} with $\tilde{\mbf}^{t+1}$, so that the generated image still obeys the mask conditioning.
Hence, for the second stage training, instead of $\loss{bg}$, we utilize $\loss{bg}'$ where
\begin{equation}
\begin{split}
    \loss{bg}'&=
    \left\|
        (1-\bm^t)\odot\tilde{\mbf}^{t+1} - (1-\bm^t)\odot\mbf^{t+1}
    \right\|_1
\end{split}
    ,
    \label{eq:loss_bg2}
\end{equation}
which now enforces our fine-tuned generator $\GenNet$ to still obey the provided mask.

%% file: fig/G_item.tex
\begin{figure*}[t]
    \centering
    \includegraphics[width=0.9\linewidth]{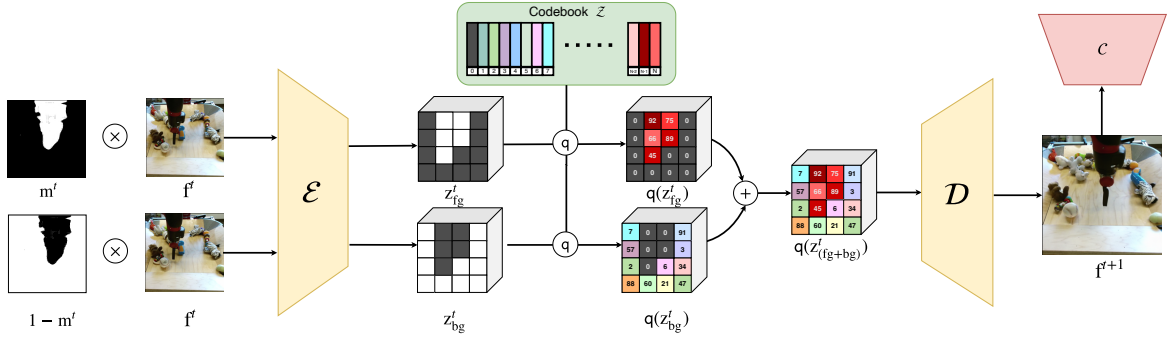}
    \caption{{\bf Frame Generator $\mathcal{G}$.} 
    We employ VQGAN framework for the generator $\mathcal{G}$, comprising of an encoder $\mathcal{E}$, decoder $\mathcal{D}$, a learnable discrete codebook $\mathcal{Z}$ and a discriminator $\mathcal{C}$. We encode the foreground and the background layers separately and then merge them in the latent space before feeding them to the decoder $\mathcal{D}$ which generates the next frame in the sequence.
    }
    \label{fig:G} 
\end{figure*}

%% file: sec/4_results.tex
\section{Experiments}
\label{sec:sec4}

While our method is {\em not} limited to a ``single-agent'' assumption, \ie, single dominant moving agent in the scene, previous work, and notably \cite{playable} which is the closest and the most competitive baseline, are.
Hence, for fair comparison, we adopt the single-agent setup for the majority of our experiments.
We train / test on the following datasets:

\paragraph{BAIR Robot Pushing Dataset \cite{bair}} 
This dataset contains 44K video clips ($256 \times 256$ resolution) of a single robot arm agent pushing toys on a flat surface. 

\paragraph{Tennis Dataset \cite{playable}} This dataset contains 900 clips extracted from two full tennis matches on YouTube.
These clips are cropped such that only the half of the court is visible.
The resolution of each frame is ($96 \times 256$).

\subsection{Results}
\label{sec:results}

\vspace{-\customparskip}
\paragraph{Evaluation Protocol}
We compare our model against other conditional generative methods, focusing on quality of reconstructed sequences and controlability. 
We evaluate our model under three control protocols: two parametric (position, affine) and one non-parametric (direct non-differentiable control over mask):
\vspace{-0.06in}
\begin{itemize}[leftmargin=*, noitemsep]
    \item Ours /w \texttt{position} control: We first use our trained mask network $\mathcal{M}$ to extract masks from the ground truth test sequences, then we use Eq.~(\ref{eq:control}) to approximate those masks with control for generation, where $\theta$ is restricted to \texttt{positional} parameters, \ie, $x$ and $y$ translation.
    \item Ours /w \texttt{affine} control: Similar to above, here our $\theta$ employs full \texttt{affine} transformation parameters, \ie, translation, rotation, scaling and shearing.
    \item Ours /w \texttt{non-param} control: We use masks predicted from ground truth test sequences themselves to condition our generation.
    These masks can change at a pixel-level, hence constituting non-parametric control. 
\end{itemize}
\vspace{-0.06in}
For testing, we generate video sequences conditioned on the first frame $f_0$ and the user input $\Theta = \{\theta^t_u\}_{t=1}^T$ in all cases.

\paragraph{Metrics} To quantitatively evaluate our results, we consider standard metrics:
\vspace{-0.06in}
\begin{itemize}[leftmargin=*, noitemsep]
  \item \textit{Learned Perceptual Image Patch Similarity (LPIPS)}~\cite{LPIPS}: LPIPS measures the perceptual distance between generated and ground truth frames. 
  \item \textit{Fréchet Inception Distance (FID)}~\cite{FID}: FID calculates the Fréchet distance between multivariate Gaussians fitted to the feature space of the Inception-v3 network of generated and ground truth frames.
  \item \textit{Fréchet Video Distance (FVD)}~\cite{FVD}: FVD extends FID to the video domain. In addition to the quality of each frame, FVD also evaluates the temporal coherence between generated and ground truth sequences.
  \item \textit{Average Detection Distance (ADD)}~\cite{playable}: ADD first uses Faster-RCNN~\cite{fasterrcnn} to detect the target object in both generated and ground truth frames, then calculates the Euclidean distance between the bound box centers. 
  \item \textit{Missing Detection Rate (MDR)}~\cite{playable}: MDR reports percentage of unsuccessful detections in generated vs. successful detections in ground truth sequences. 
  \item \textit{Rooted Mean Square Error of Displacement (RMSED)}~\cite{playable}: RMSED, which we define, reports the RMSE of the displacement of ground truth locations vs. generated locations. See Figure~\ref{fig:compare_bair} for more details.
\end{itemize}
LIPIPS, FID and FVD measure the quality of generated videos. ADD and MDR measure how the action label conditions the generated video, and RMSED measures the precision of control.

\paragraph{Baselines}
CADDY \cite{playable} is the only unsupervised video generation method that allows frame-level user conditioning, thus we use it as our main baseline.
We also include results of other frame-level conditioned methods: MoCoGAN \cite{Tulyakov_CVPR18}, SAVP \cite{Lee_2018}, and their high-resolution adaptations MoCoGAN+ and SAVP+ from \cite{playable}.

\input{tab/bair.tex}

\input{tab/tennis.tex}

\paragraph{Quantitative Results} We report the results on the \textit{BAIR} dataset in Table~\ref{tab:bair}.
We highlight that in terms of RMSED score, our method achieved the highest precision of control (more than $\times5$ improvement compared to other baselines). In terms of generation quality,  with similar level of abstraction of ground truth information (ours: 6 continuous \texttt{affine} control parameters, CADDY: 7 discrete action labels), our model outperformed CADDY on all three evaluated metrics by a large margin, demonstrating that our model is of better generation quality. With \texttt{non-param} control, our generation quality is comparable to the SAVP+. 

Table~\ref{tab:tennis} shows the results on the \textit{Tennis} dataset. In terms of generation quality (LPIPS, FID, FVD), all our adaptations outperformed all other comparing methods.
Specifically, in terms of FID score, our model is up to 37\% better than the closest baseline (\cite{playable}).
Further, in terms of control precision, our method achieves the lowest error on ADD and MDR (improvement of 80\% \& 70\% respectively), indicating our method is able to generate consistent players with accurate control.
One can see that simple positional control works much better here compared to the \textit{BAIR} dataset. This can be attributed to largely in plane motion of the subject.

\input{fig/compare_bair}
\input{fig/compare_tennis}
\input{fig/control_item}

\paragraph{Qualitative Results}
In Figure~\ref{fig:compare_bair} and Figure~\ref{fig:compare_tennis}, we show generated sequences on the \textit{BAIR} and \textit{Tennis} dataset (we used our model with \texttt{affine} control for both cases shown). In terms of image quality, our method is superior to competitors. In terms of control accuracy, unlike other competing methods, our method is able to precisely place the robot arm and the tennis player in the correct position. 

In Figure~\ref{fig:control}, we show the results of our model reacting to different user control signals. On the \textit{Tennis} dataset, our method not only moves the player in the correct direction, it's also able to generate plausible motions of the player itself. On the \textit{BAIR} dataset, our model is able to ``hullcinate'' what's missing in the original frame and generate frames with respect to the control signal, \ie, in the ``down, 35 pixels'' example, our model successfully generates the upper part of the robot arm, not available on the input frame.

\input{fig/fancy_item}

In Figure~\ref{fig:fancy}, we show that our method is capable of generating and controlling videos with multiple moving objects by simply overlaying two individually controlled mask sequences together, \ie, producing 2 and 3 players in this example. As far as we know, our method is the only video generation method that allows frame-level control of multiple objects acting in the same scene. We provide more visual results in the \supplementary

\subsection{Ablation Study}
\label{sec:ablation}

\input{fig/ablation}
\textbf{Mask losses.}
Here we explore impact of our 
key design choices have on the quality of generated results and the foreground mask. 
We show quantitative and qualitative results in Figure ~\ref{fig:ablation_mask}. 
The background loss $\loss{\text{bg}}$ enforces the network to generate meaningful masks, without it, the mask network fails to generate a reasonable mask (all zeros). 
The foreground constraint $\loss{\text{fg}}$ shrinks the mask as much as possible. Without this term the network learns a travail solution, where $\loss{\text{bg}}$ in Eq.(\ref{eq:loss_bg}) becomes 0 -- labeling everything as the foreground (all ones).
When computing the foreground loss $\loss{\text{fg}}$, we introduce a dynamic mask size prior. We ablate this choice by instead using a fixed global prior of $0.15$ as in \cite{Bielski_NeurIPS2019}. 
Visuals show that if we do not use dynamic prior, the network tends to generate masks with a fixed size, which leads to hollow masks for samples with larger foreground. 
To prevent information leaking from soft masks, we binarize the masks with thresholding the mask value, without the binary loss $\loss{\text{bin}}$, some pixels on the mask fails to pass the threshold and leave some defects on the binarized mask.
Overall, the ablations show that all our design choices are important. %

\paragraph{One-stage training VS. Two-stage training}
\label{sec:ablation}
Breaking our training procedure into two stages is a crucial design for the performance of our method. As described in \Eq{optimization}, a well-trained mask generator is a prerequisite for finding the \textit{pseudo user control $\hat\btheta$}, which we use to introduce controllability to our model. Nevertheless, we still experimented with training the model with one single shot (training Stage II directly by replacing $\mbf^t$ with $\mbf^{t+1}$).
This leads to vastly poorer performance during test time (Figure \ref{fig:ablation_mask}, ``single-stage training").

%% file: tab/bair.tex
\begin{table}
\centering
\resizebox{\linewidth}{!}{ %

\begin{tabular}{@{}lllll@{}}
\toprule
Method          & LPIPS $\downarrow$& FID $\downarrow$  & FVD $\downarrow$ & RMSED $\downarrow$ \\
\midrule
MoCoGAN \cite{Tulyakov_CVPR18}       & 0.466 & 198 & 1380 & - \\
MoCoGAN+ (from \cite{playable})      & 0.201 & 66.1 & 849 & 0.211 \\
SAVP \cite{Lee_2018}        & 0.433 & 220 & 1720 & - \\
SAVP+ (from \cite{playable}) & \textbf{0.154} & \textbf{27.2} & 303 & 0.109 \\
\midrule
CADDY \cite{playable} & 0.202 & 35.9 & 423 & 0.132 \\
Ours /w {\tt position} control           & 0.202 & 28.5 & 333 & 0.059           \\
Ours /w {\tt affine} control           & 0.201 & 30.1 & \textbf{292} & 0.035          \\
Ours /w {\tt non-param} control        & 0.176 & 29.3 & 293 & \textbf{0.021}            \\
\bottomrule
\end{tabular}
}
\caption{Results on the \textit{BAIR} Dataset}
\label{tab:bair}
\end{table}

%% file: tab/tennis.tex
\begin{table}
\resizebox{\linewidth}{!}{ %

\begin{tabular}{@{}llllll@{}}
\toprule
Method  & LPIPS $\downarrow$ & FID $\downarrow$ & FVD $\downarrow$ & ADD $\downarrow$ & MDR $\downarrow$  \\
\midrule
MoCoGAN \cite{Tulyakov_CVPR18}  & 0.266 & 132  & 3400 & 28.5 & 20.2 \\
MoCoGAN+ (from \cite{playable}) & 0.166 & 56.8 & 1410 & 48.2 & 27.0  \\
SAVP \cite{Lee_2018}            & 0.245 & 156  & 3270 & 10.7 & 19.7\\
SAVP+ (from \cite{playable})    & 0.104 & 25.2 & 223  & 13.4 & 19.2  \\
\midrule
CADDY \cite{playable}           & 0.102 & 13.7 & 239  & 8.85 & 1.01  \\
Ours /w {\tt position} control   & 0.122 & 10.1 & 215  & 4.30 & \textbf{0.300} \\
Ours /w {\tt affine} control     & 0.115 & 11.2 & 207 & 3.40 & 0.317 \\
Ours /w {\tt non-param} control  & \textbf{0.100} & \textbf{8.68} & \textbf{204}  & \textbf{1.76} & 0.306 \\
\bottomrule
\end{tabular}
}
\caption{Results on the \textit{Tennis} Dataset}
\label{tab:tennis}
\end{table}

%% file: fig/compare_bair.tex
\begin{figure}
    \centering
    \includegraphics[width=0.9\linewidth]{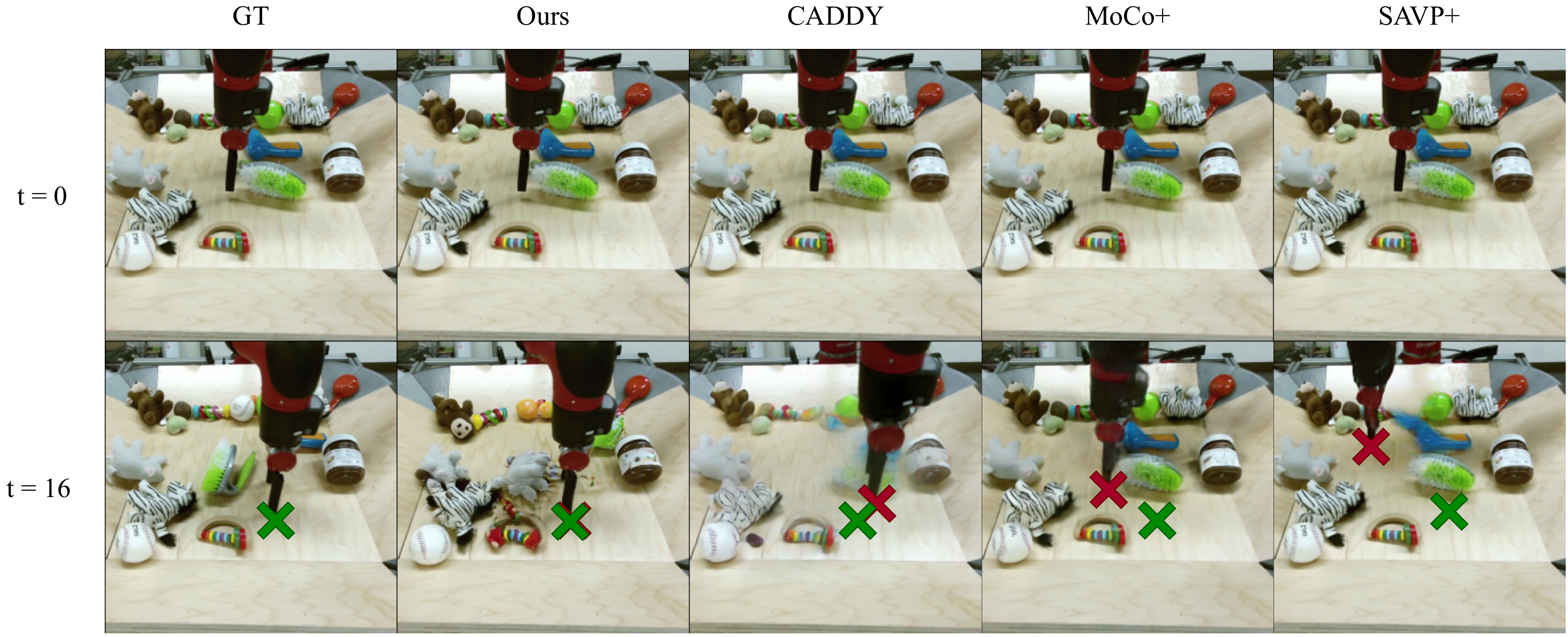}
    \caption{{\bf RMSED.} 
    We labelled the robot arm positions for half of the testing sequences (128 videos out of 256) from both generated videos and ground truth videos. As illustrated on the figure: GT location is marked {\color{dark_green}green} and generated locations are marked {\color{red}red}. We calculate RMSE of the displacement between GT locations and generated locations to arrive at the RMSED score.
    }
    \label{fig:compare_bair} 
\end{figure}

%% file: fig/compare_tennis.tex
\begin{figure}
    \centering
    \includegraphics[width=1\linewidth]{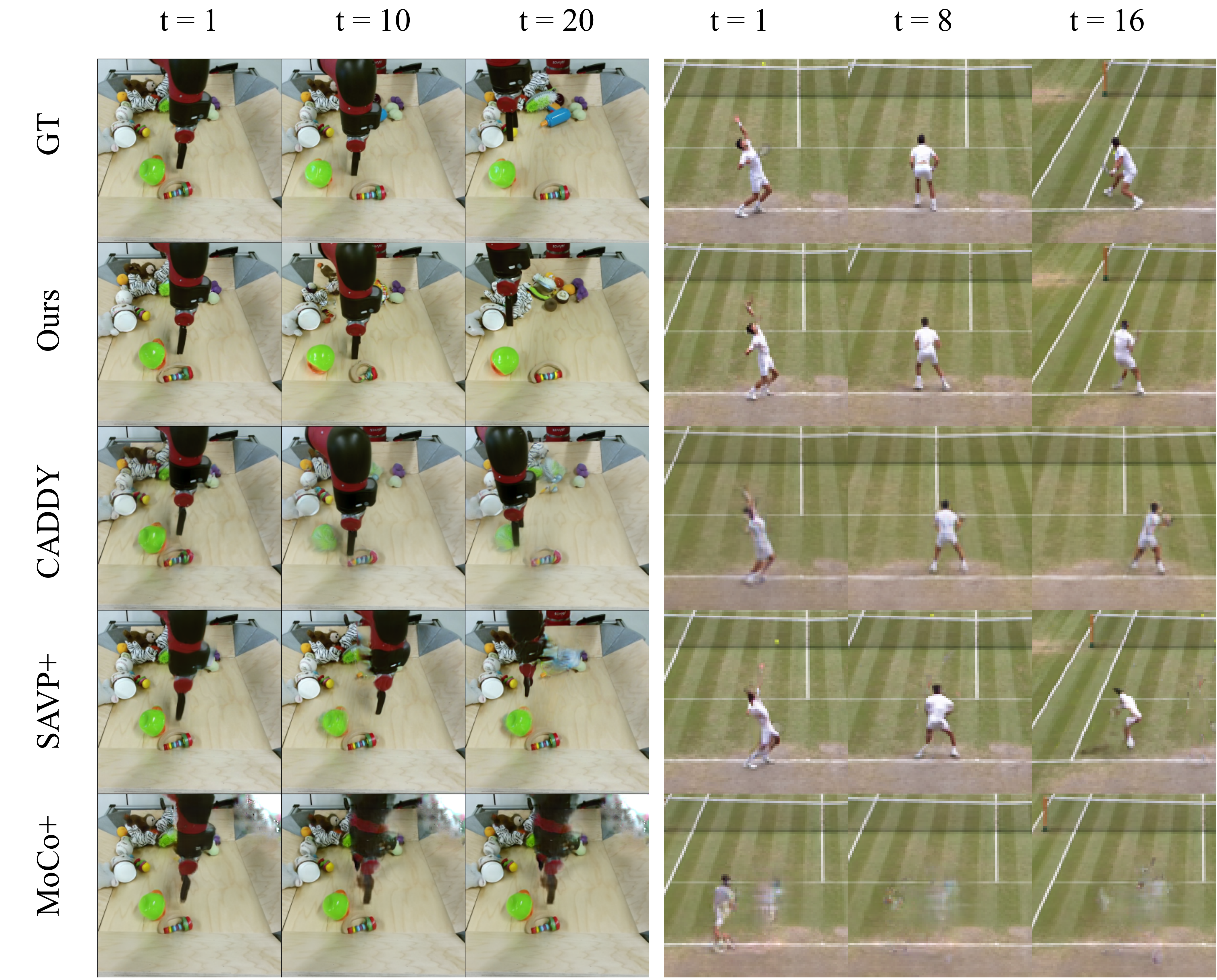}
    \caption{{\bf Qualitative Results.} Comparison of our model with \texttt{affine} control against other conditional generative methods. \textit{Tennis} sequences are cropped for better visualization. }
    \label{fig:compare_tennis}
\end{figure}

%% file: fig/control_item.tex
\begin{figure*}
    \centering
    \includegraphics[ width=0.9\linewidth,trim=0 50 0 0]{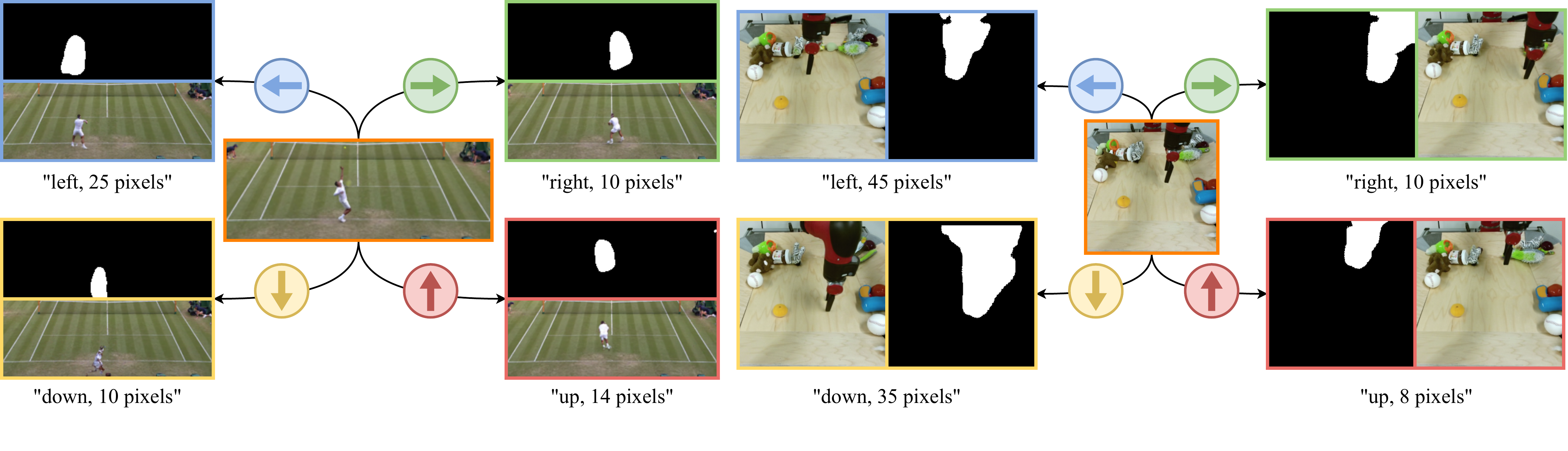}
    \caption{{\bf Effectiveness of Control.} Illustrated is how our model precisely reacts to different controlling signals starting from the same initial frame. We illustrate position parameters; however, other affine control parameters are also possible (\eg, scale, rotation and shear).}
    \label{fig:control}
\end{figure*}

%% file: fig/fancy_item.tex
\begin{figure}

    \centering
    \includegraphics[width=1\linewidth]{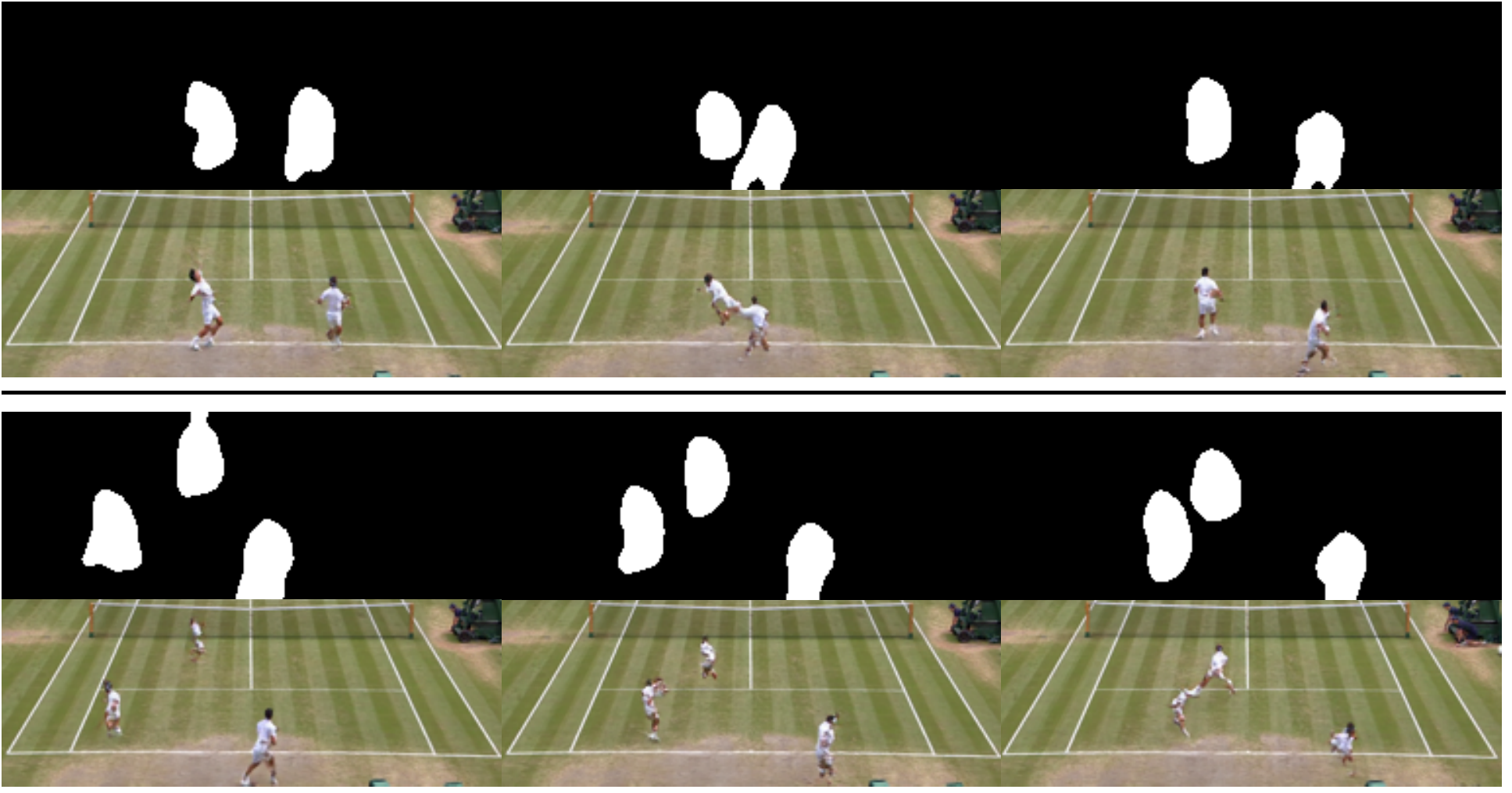}
    \caption{{\bf Control of Multiple Agents}. Our method is able to generate videos with multiple moving objects that can be controlled individually by their respective masks.}
    \label{fig:fancy}

\end{figure}

%% file: fig/ablation.tex
\begin{figure}
\begin{minipage}{\linewidth}
    \centering

    \resizebox{\linewidth}{!}{
        \setlength{\tabcolsep}{10pt}
        \begin{tabular}{@{}llll@{}}
        \toprule
        Method          & LPIPS $\downarrow$& FID $\downarrow$  & FVD $\downarrow$ \\
        \midrule
        w/o $\mathcal{L}_{fg}$        & 0.333 & 60.1 & 816 \\
        w/o $\mathcal{L}_{bg}$        & 0.306 & 97.1 & 796 \\
        w/o $\mathcal{L}_{bin}$        & 0.222 & 59.2 & 398 \\
        w/o mask prior        & 0.208 & 55.0 & \textbf{279} \\
        single-stage training        & 0.608 & 302.3 & 6614 \\
        \textbf{full}                        & \textbf{0.176} & \textbf{29.3} & 293             \\
        \bottomrule
        \end{tabular}
    }
    \includegraphics[width=\linewidth,trim=0 0 0 5]{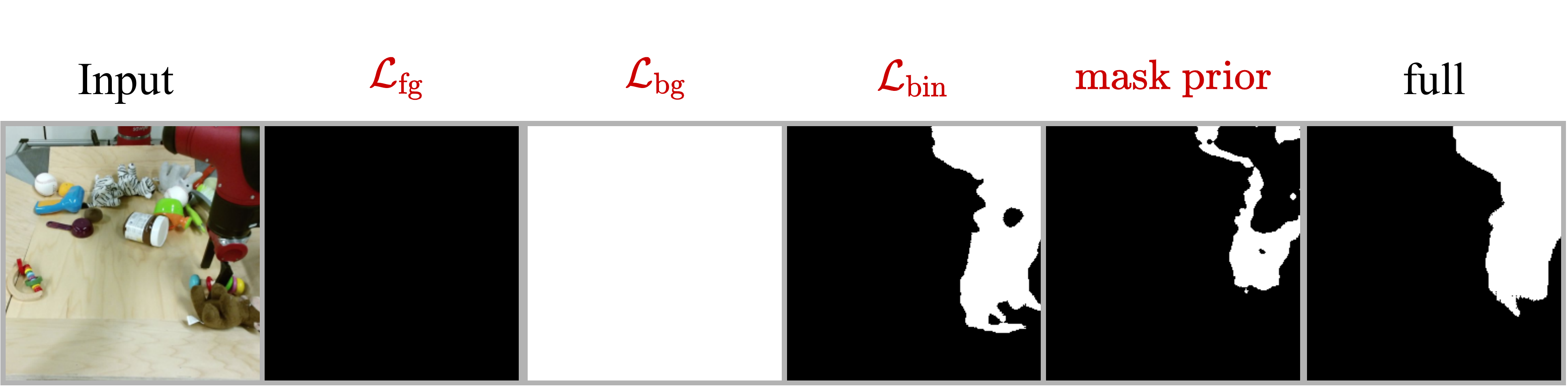}
    \caption{{\bf Ablation of Design Choices.} We \textcolor{red}{ablate} various loss terms, the use of dynamic mask prior and the two-stage training design. Thumbnails below illustrate the effect these components have on the estimated mask itself; full model producing the most coherent mask.} 
    \label{fig:ablation_mask}
\end{minipage}
\end{figure}

%% file: sec/5_conclusions.tex
\section{Conclusions}
\vspace{-0.02in}

We have introduced layered controllable video generation, an unsupervised method that decomposes frames into foreground and background, with which the user can control the generative process at a frame-level by altering the foreground mask.
Our core contributions are the framework itself, and the two-stage training strategy that allows our model to learn to both separate and control on its own.
We show that various degrees of control can be implemented with our method, from parametric (position, affine) to complete non-parametric
control with the mask.
Our results on \textit{BAIR} and \textit{Tennis} datasets show that our method outperforms the state-of-the-art in both quality and control.

%% file: sec/6_acknowledgements.tex
\section{Acknowledgements}
\vspace{-0.02in}
This work was funded, in part, by the Vector Institute for AI, Canada CIFAR AI Chair, NSERC CRC and an NSERC Discovery and Discovery Accelerator Grants. Resources used in preparing this research were provided, in part, by the Province of Ontario, the Government of Canada through CIFAR, and companies sponsoring the Vector Institute \url{www.vectorinstitute.ai/#partners}. Additional hardware support was provided by John R. Evans Leaders Fund CFI grant and Compute Canada under the Resource Allocation Competition awards of the two PIs. We would also like to thank Willi Menapace for his help in answering our questions and helping with fair comparisons to \cite{playable}.

%% file: sec/X_supplementary.tex
\appendix

\setcounter{page}{1}

{
\centering
\Large
\textbf{Layered Controllable Video Generation} \\
\vspace{0.5em}Supplementary Material \\
\vspace{1.0em}
}

\appendix
This supplementary material is structured as follows.
In \Section{A}, we present our architectural details of our models, and the implementation details of our training/ testing procedures.
In \Section{B}, we show visual results of the frames generated by our model under different conditions.
In \Section{C}, we show additional ablation results to prove the effectiveness of our two staged training setup. 
In \Section{D}, we show soft masks could lead to degenerated results.
Our video results can be found at 
\href{https://gabriel-huang.github.io/layered_controllable_video_generation/}{this website}.
\section{Implementation Details}
\label{sec:A}
\paragraph{Architecture details}
Our model consists of four main modules: the \textit{Mask Network} $\mathcal{M}$, the \textit{Encoder} $\mathcal{E}$, the learnable discrete code book $\mathcal{Z}$, and the \textit{Decoder} $\mathcal{D}$. 
We summarize the details in \Table{architecture}.
In total, our model has $47.8$ M parameters, and takes about $191.2$ MB to store on disk.
\input{tab/architecture}

\paragraph{Training Details}
In all our experiments, we use the Adam Optimizer with a fixed learning rate of $4.5\times 10^{-6}$, which we found empirically.
\begin{itemize}[leftmargin=*, noitemsep]
    \item \textit{Initial training for mask-based generation:}
    For this stage, we train all models for $10,000$ iterations.
    For experiments on the \textit{BAIR} dataset, we initialize the ratio between $\lambda_{\text{bg}} : \lambda_{\text{fg}}$ to be $80 : 1$, and reduce it by a factor of $2$ for every $1,000$ iterations until this ratio becomes $5:1$.
    For experiments on the \textit{Tennis} dataset, this ratio is initialized to  $10:1$ and reduced by half after $5,000$ iterations.
    
    \item \textit{Fine-tuning for controllability:} In this stage of training, we drop all losses related to the mask regularization as their gradient becomes zero (see \Section{phaseII}), except for $\mathcal{L}_\text{bg}$. 
    On the \textit{BAIR} dataset, $\lambda_{\text{bg}}$ is set to $5$, and on the \textit{Tennis} dataset, it's set to $0.5$.
    
    \item \textit{Finding the ``ground-truth'' control signal:} In Equation.~\ref{eq:control}, we use the fully differentiable transformation operation $\mathcal{T}(\theta^t)$ to approximate the mask of the  ground truth next frame using the mask of the current frame.
    For each frame, we optimize the affine transformation matrix that's been used to transform $\bm^t$ into $\bm_c^t$, the optimization is performed by the ADAM optimizer for $1,000$ iterations with the learning rate of $0.1$.
    
On the \textit{BAIR} dataset, our model requires $80$GB GPU memory and $40$ hours to train on $4\times$ RTX-6000 (24GB V-RAM per GPU) GPUs, and on the \textit{Tennis} daaset, the model takes $22$GB GPU memory and $23$ hours to train on a single RTX-3090 (24GB V-RAM).
\end{itemize}

\section{Additional qualitative results}
\label{sec:B}

\begin{figure*}
    \centering
    \includegraphics[width=1\linewidth]{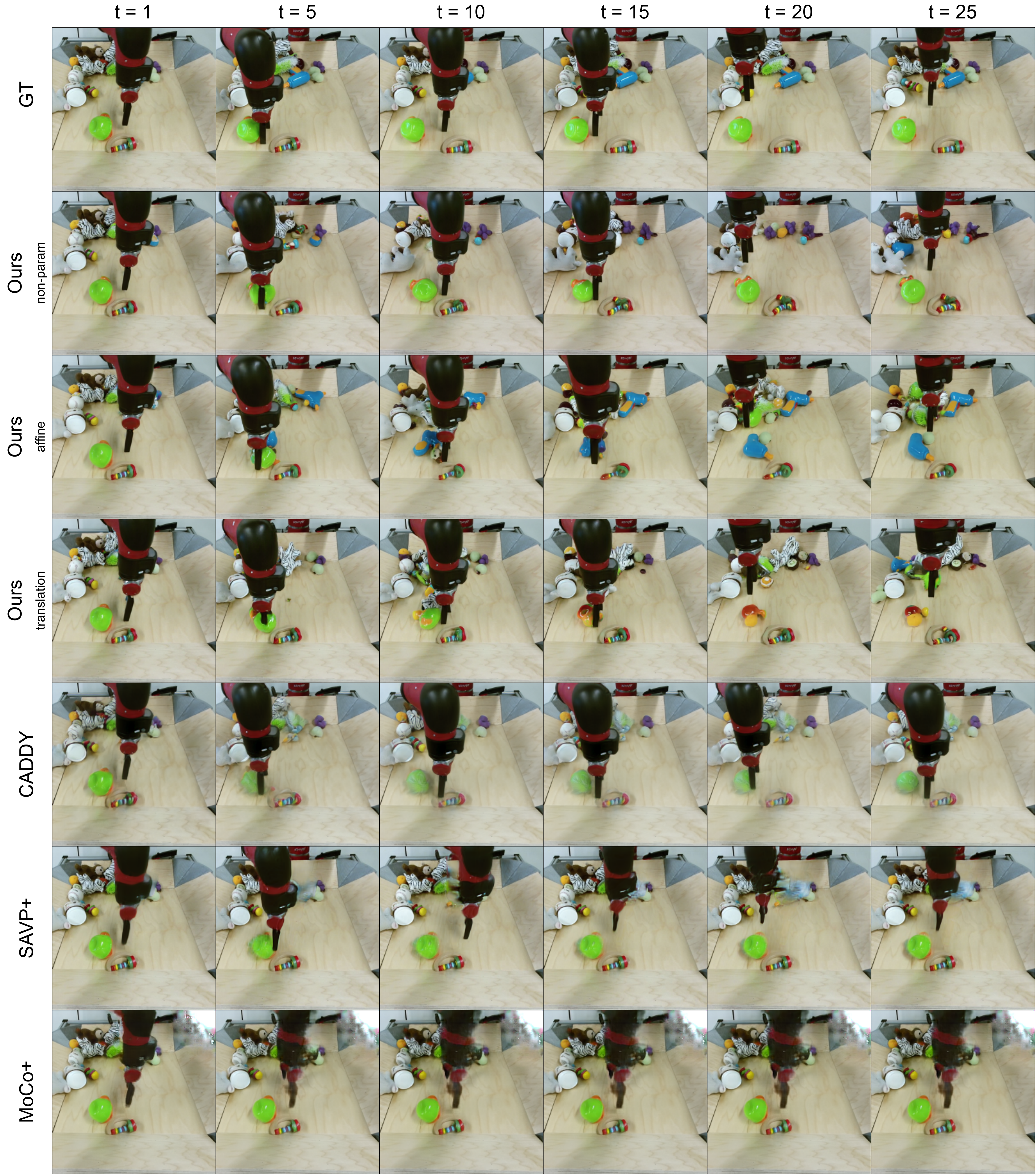}
    \caption{
        Comparing with other baselines on the \textit{BAIR} dataset.
    }
    \label{fig:bair_compare} 
\end{figure*}

\begin{figure*}
    \centering
    \includegraphics[width=1\linewidth]{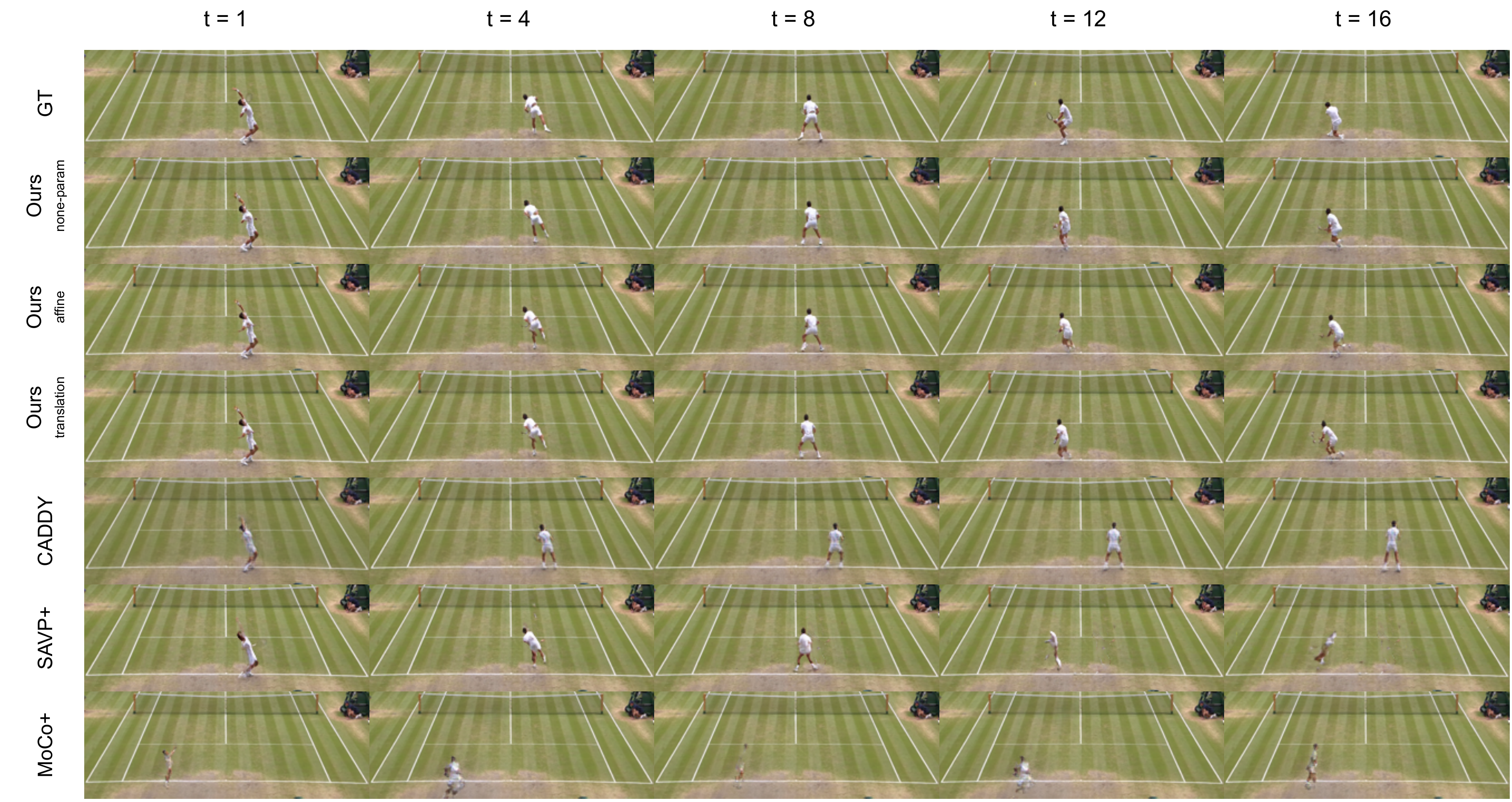}
    \caption{
        Comparing with other baselines on the \textit{Tennis} dataset.
    }
    \label{fig:tennis_compare} 
\end{figure*}

\paragraph{Comparing with other baselines} 
In \Figure{bair_compare} and \Figure{tennis_compare}, we show the comparison of the three variants of our model against other baselines that we mentioned in the main text, on the \textit{BAIR} and \textit{Tennis} datasets respectively.
Noticeably on both datsets, our model achieves the most precise control over the generated video (see how the generated motion corresponds to the ground truth sequences), meanwhile generating one of the best frame-quality videos. 
For example, CADDY fails to control the robot arm, SAVP+ controls it up to some degree, but as shown in frames $t{=}10$ and $t{=}15$, their control is not as precise, whereas all of our variants provide highly accurate control.

\begin{figure*}
    \centering
    \includegraphics[width=1\linewidth]{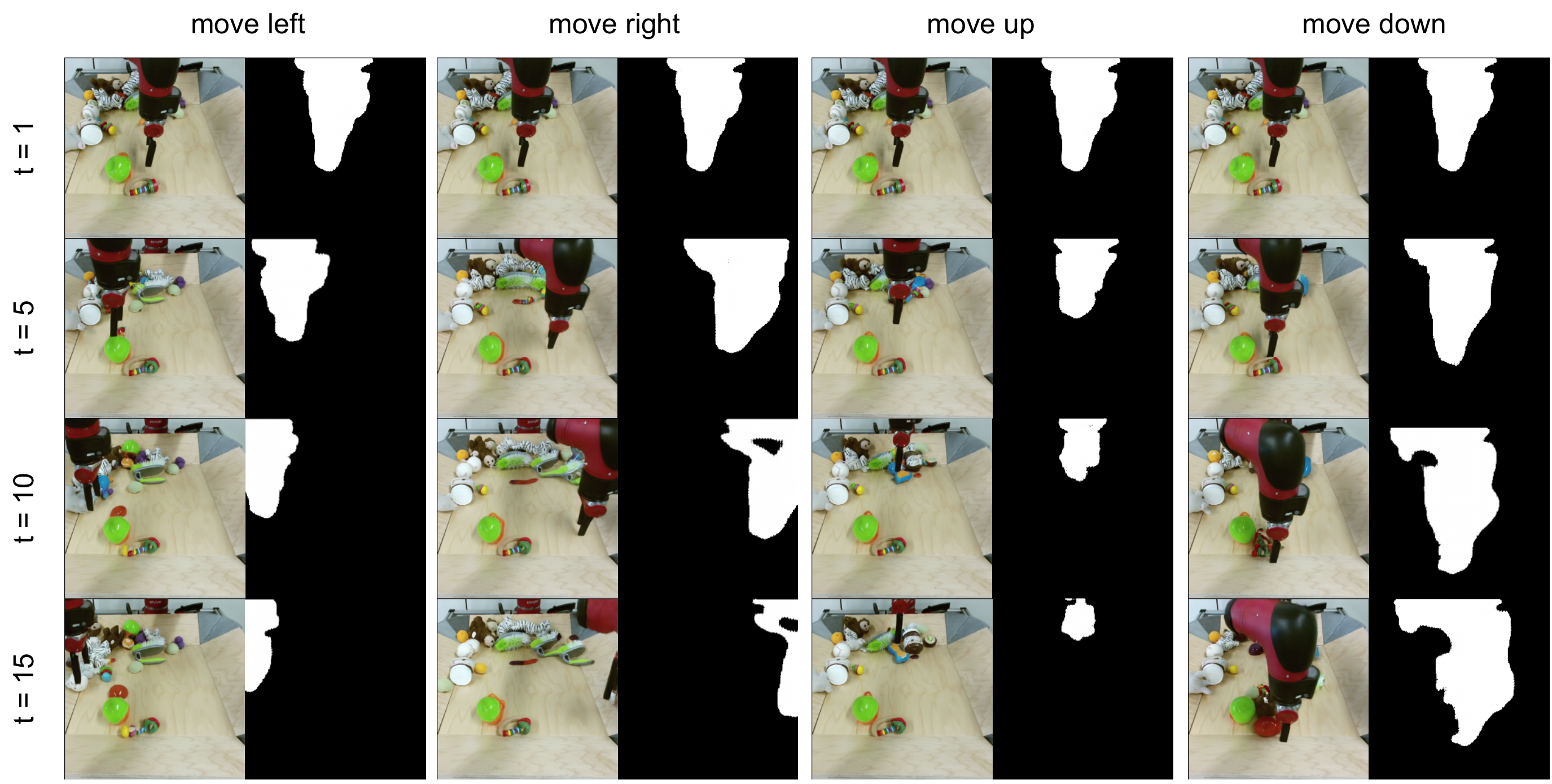}
    \caption{
        Generating videos conditioned on user inputs, \textit{BAIR} dataset.
    }
    \label{fig:control_test_bair} 
\end{figure*}

\begin{figure*}
    \centering
    \includegraphics[width=1\linewidth]{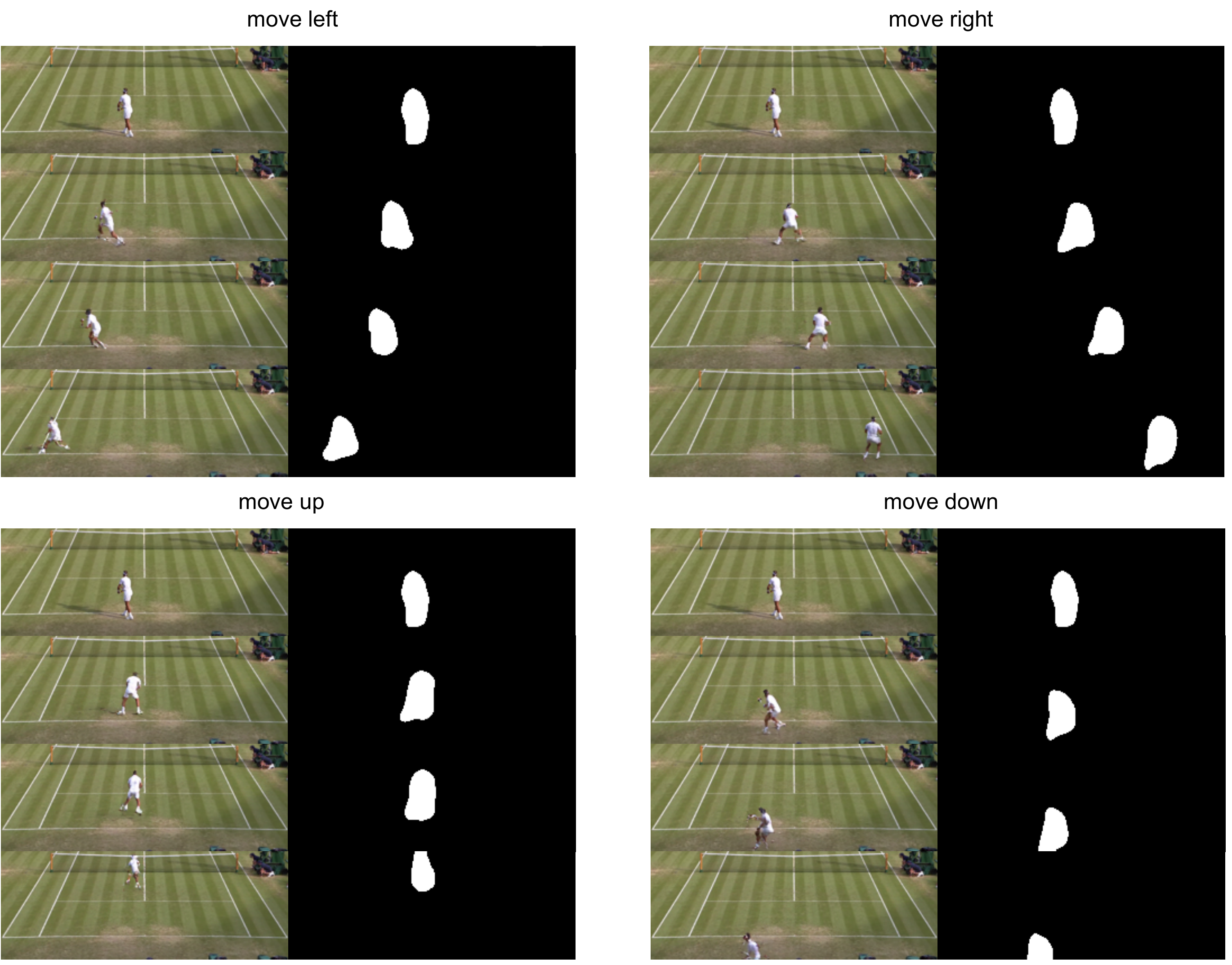}
    \caption{
        Generating videos conditioned on user inputs, \textit{Tennis} dataset.
    }
    \label{fig:control_test_tennis} 
\end{figure*}

\paragraph{Generating videos conditioned on user inputs}
In \Figure{control_test_bair} and \Figure{control_test_tennis}, we show how the video generated by our model responds to the user inputs, on the \textit{BAIR} and \textit{Tennis} datasets respectively. Here, for each sample, we keep applying the same control signal (moving left/right/up/down) to the model so that it can generate video sequences with single consistent motions.

\begin{figure*}[t!]
    \centering
    \includegraphics[width=1\linewidth]{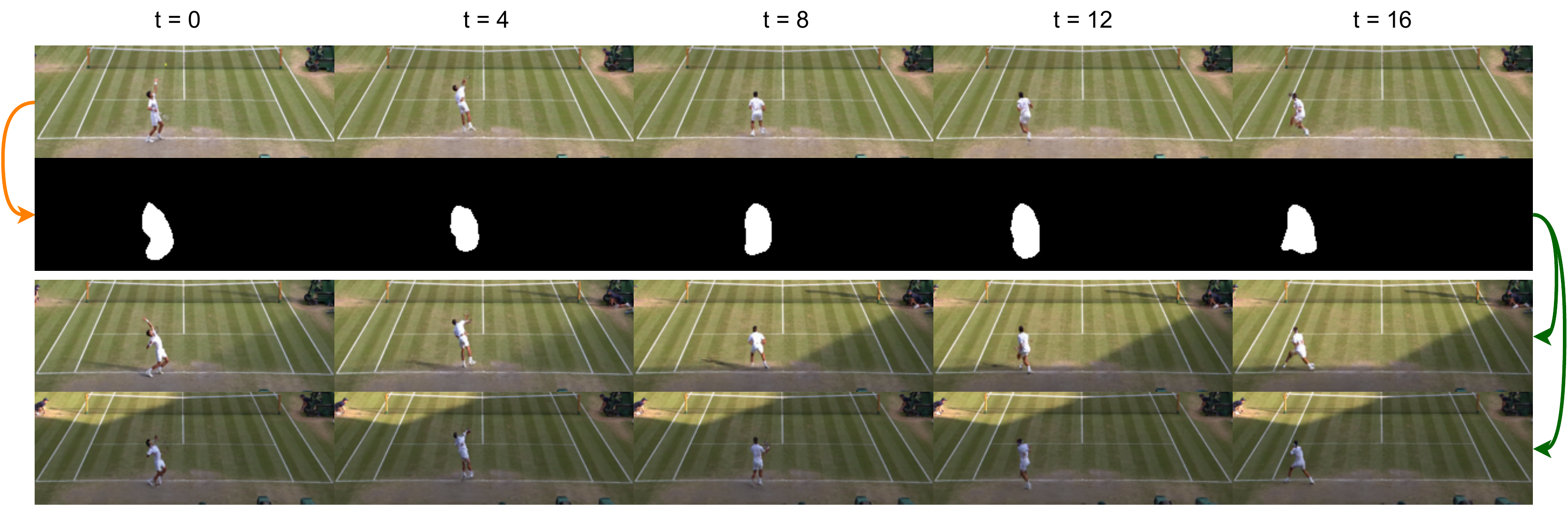}
    \caption{
        ``Mimicking''. Our model can be used to extract the motion from a driving sequence and then apply onto different appearances (staring frames). The mask sequence is extracted from the driving video (first row), then applied onto two other staring frames (third, fourth row).
    }
    \label{fig:mimic} 
\end{figure*}

\begin{figure*}
    \centering
    \includegraphics[width=1\linewidth]{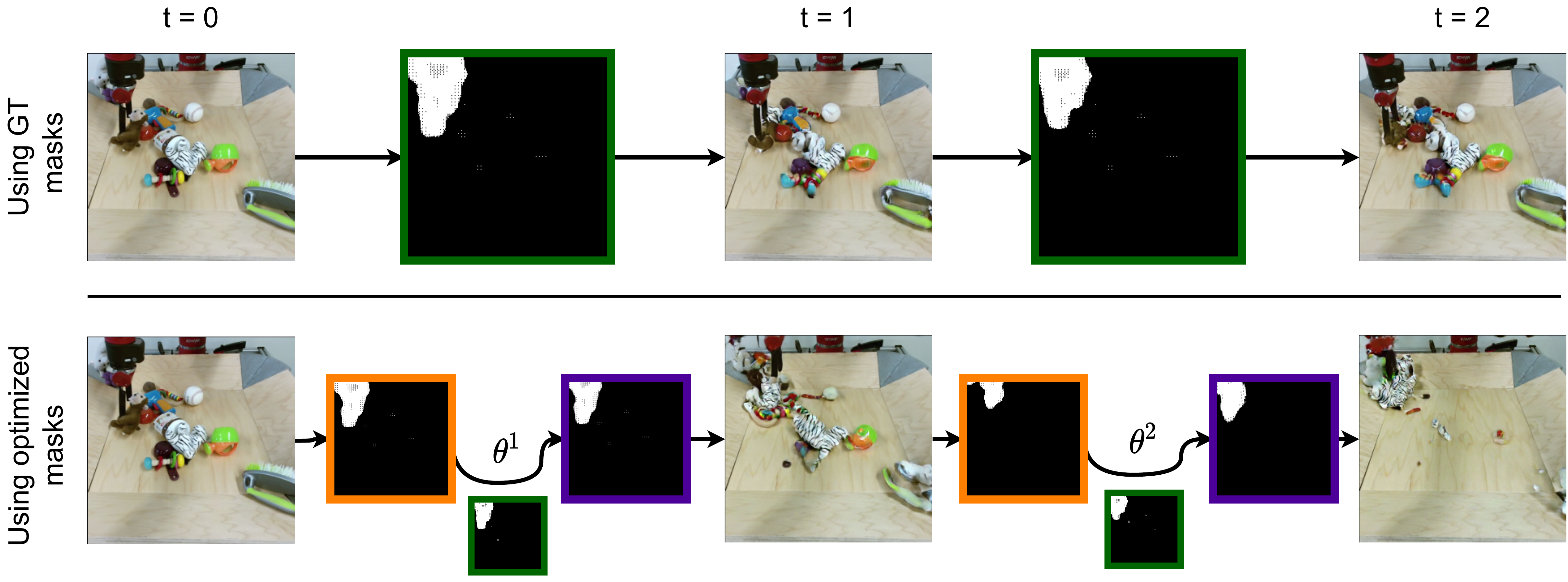}
    \caption{
        Single stage training results in poor testing performance when control is introduced. {\color{dark_green}Green} represents GT next frame mask, {\color{orange_bright}orange} represents generated current frame's mask, and {\color{purple}purple} represents transformed mask. On the top row, the model is tested using ground truth masks to generate future frames (same as how the model was trained). On the bottom row, the model is tested using pseudo control signals (same as real-world testing). The performance of the single stage trained model drops significantly when there are edits to the mask.
    }
    \label{fig:onestage} 
\end{figure*}

\begin{figure*}
    \centering
    \includegraphics[width=1\linewidth]{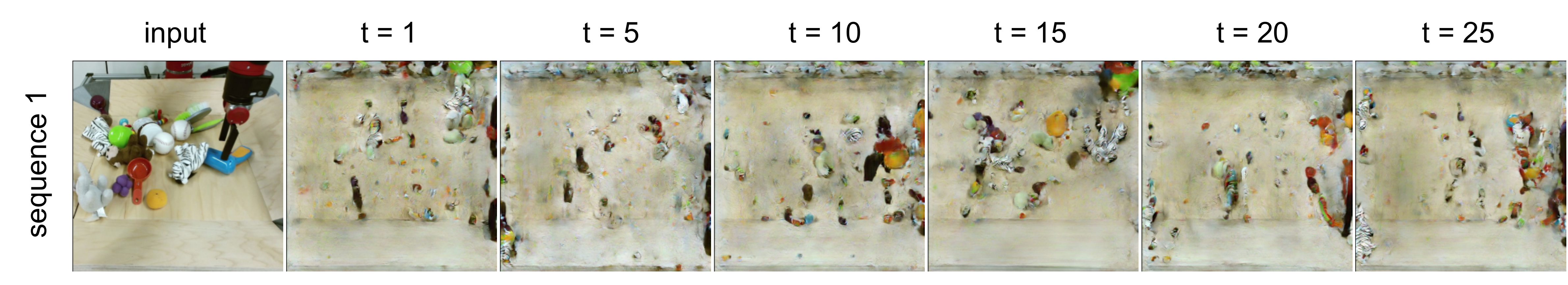}
    \caption{Soft masks result in degenerated frames.}
    \label{fig:softmask}
\end{figure*}

\paragraph{Action mimicking}
In \Figure{mimic}, we show that our method can be used to extract the motion from a driving sequence and then apply onto different appearances (staring frames).

\makeatletter
\setlength{\@fptop}{0pt}
\makeatother

\paragraph{Video results} 
For more video results, please refer to \href{https://gabriel-huang.github.io/layered_controllable_video_generation/}{this website}.
, where we demonstrate all above mentioned results in video format, as well as other applications of our method, such as generating videos of multi players, real-time user controllable generation demo, and animating a single frame with different motions.

\section{Additional Ablation results - Single Stage Training}
\label{sec:C}
As mentioned in our main text, breaking down our training procedure into two stages is a crucial design in our setup. In stage I, our model focuses on generating a FG/BG segmentation mask, and in stage II, we make edits to this mask to finetune the network for controlled generation. 
Training the model in one single stage wouldn't allow us to introduce controllability, without stage II finetuning, the model is extremely vulnerable to any changes to the mask. As shown on Figure \ref{fig:onestage}, on the top row, the model is tested using ground truth masks to generate future frames (same as single stage training setup), and the model performed well. However, on the bottom row, when we try to use pseudo control signals to transform the mask, then use the transformed mask to condition the generation (same as real-world testing for controlled video generation), the model performs poorly.

\section{Additional Ablation results - Soft Masks}
\label{sec:D}
As shown in Fig.~\ref{fig:softmask}, without mask binarization, information is leaked through the soft masks, shifting the mask will not only affect the foreground, but the entire scene, which results in degeneration of the frame.

%% file: tab/architecture.tex
\begin{table}[t!]
\centering
\caption{Architecture details of the main modules of our model. The high-level design follows the architecture presented in \cite{taming}. $K$ denotes the number of entries in the code book, $n_z$ is the dimension of each entry, $h,w = (H, W)/4$ and $h^\prime, w^\prime = (H,W)/16$. For the discriminator $\mathcal{C}$, we use the vanilla PatchGAN discriminator as described in \cite{pix2pix}.}
\begin{tabular}{c}
\toprule
Mask Network $\mathcal{M}$ \\      
\midrule
$x \in \mathbb{R}^{H \times W \times 3}$ \\
Conv2D $\rightarrow \mathbb{R}^{H \times W \times 64}$ \\
$2 \times$\{Residual Blocks + Downsample\} $\rightarrow \mathbb{R}^{h \times w \times 256}$ \\
$9 \times \text{Residual Blocks} \rightarrow \mathbb{R}^{h \times w \times 256}$ \\
$2 \times$\{Upsample + Residual Blocks \}$ \rightarrow \mathbb{R}^{h \times w \times 64}$ \\
Conv2D + Sigmoid$ \rightarrow \mathbb{R}^{H \times W \times 1}$ \\
\midrule
Encoder $\mathcal{E}$\\
\midrule
$x \in \mathbb{R}^{H \times W \times 3}$ \\
Conv2D $\rightarrow \mathbb{R}^{H \times W \times 256}$ \\
$4 \times$\{Residual Blocks + DownSample\} $\rightarrow \mathbb{R}^{h^\prime \times w^\prime \times 64}$ \\
Residual Block $\rightarrow \mathbb{R}^{h^\prime \times w^\prime \times 64}$ \\
Attention Block $\rightarrow \mathbb{R}^{h^\prime \times w^\prime \times 64}$ \\
Residual Block $\rightarrow \mathbb{R}^{h^\prime \times w^\prime \times 64}$ \\
GroupNorm, SiLu, Conv2D  $\rightarrow \mathbb{R}^{h^\prime \times w^\prime \times n_z}$ \\
\midrule
Code book $\Codebook$\\
\midrule
$K = 1024$ \\
$n_z = 256$ \\
\midrule
Decoder $\mathcal{D}$\\
\midrule
$\quant(\bz) \in \mathbb{R}^{h^\prime \times w^\prime \times n_z}$ \\
Conv2D $\rightarrow \mathbb{R}^{h^\prime \times w^\prime \times 64}$ \\
Residual Block $\rightarrow \mathbb{R}^{h^\prime \times w^\prime \times 64}$ \\
Attention Block $\rightarrow \mathbb{R}^{h^\prime \times w^\prime \times 64}$ \\
Residual Block $\rightarrow \mathbb{R}^{h^\prime \times w^\prime \times 64}$ \\
$4 \times$\{Residual Blocks + DownSample\} $\rightarrow \mathbb{R}^{H \times W \times 256}$ \\
GroupNorm, SiLu, Conv2D  $\rightarrow \mathbb{R}^{H \times W \times 3}$ \\
\bottomrule
\end{tabular}
\vspace{1em}

\label{tab:architecture}
\vspace{-2em}
\end{table}